\documentclass[10pt,twocolumn,letterpaper]{article}

\usepackage{cvpr}              % To produce the CAMERA-READY version
\definecolor{cvprblue}{rgb}{0.21,0.49,0.74}
\usepackage[pagebackref,breaklinks,colorlinks,allcolors=cvprblue]{hyperref}
\usepackage{stfloats}
\usepackage{multirow}
\usepackage[table]{xcolor}
\usepackage{makecell}
\usepackage{listings}
\usepackage[most]{tcolorbox}
\tcbuselibrary{listings, breakable,xparse,skins}
\definecolor{webcolor}{HTML}{90A9B6}

\usepackage{cuted}            
\usepackage{caption}   
\usepackage{placeins}
\usepackage{minted} 
\tcbuselibrary{minted}
\def\paperID{*****} % *** Enter the Paper ID here
\def\confName{CVPR}
\def\confYear{2026}
\newcommand{\website}{\href{https://chenghaogu.github.io/IGen/}{\textcolor{webcolor}{https://chenghaogu.github.io/IGen/}}}

\title{IGen: Scalable Data Generation for Robot Learning from Open-World Images}

\author{
Chenghao Gu$^{1*}$ \quad 
Haolan Kang$^{2*}$ \quad 
Junchao Lin$^{3*}$ \quad
Jinghe Wang$^1$  \quad
Duo Wu$^1$  \quad
\\
Shuzhao Xie$^1$  \quad
Fanding Huang$^1$  \quad
Junchen Ge$^1$  \quad
Ziyang Gong$^4$ \quad
Letian Li$^1$ \quad
\\
Hongying Zheng$^5$  \quad
Changwei Lv$^5$ \quad
Zhi Wang$^{1\dagger}$ \quad    
\vspace{0.2cm} \\
$ ^1$Shenzhen International Graduate School, Tsinghua University \\
$ ^2$The University of Hong Kong \quad
$ ^3$Beijing University of Chemical Technology \\
$ ^4$Shanghai Jiao Tong University \quad
$ ^5$Shenzhen University of Infomation Technology \\[0.3em]
\website
}

\newcommand{\duo}[1]{{\color{black}#1}}

\begin{document}

\twocolumn[{
\maketitle
\begin{center}
    \centering
    \captionsetup{type=figure}
    \vspace{-0.8cm}
    \includegraphics[width=0.99\linewidth]{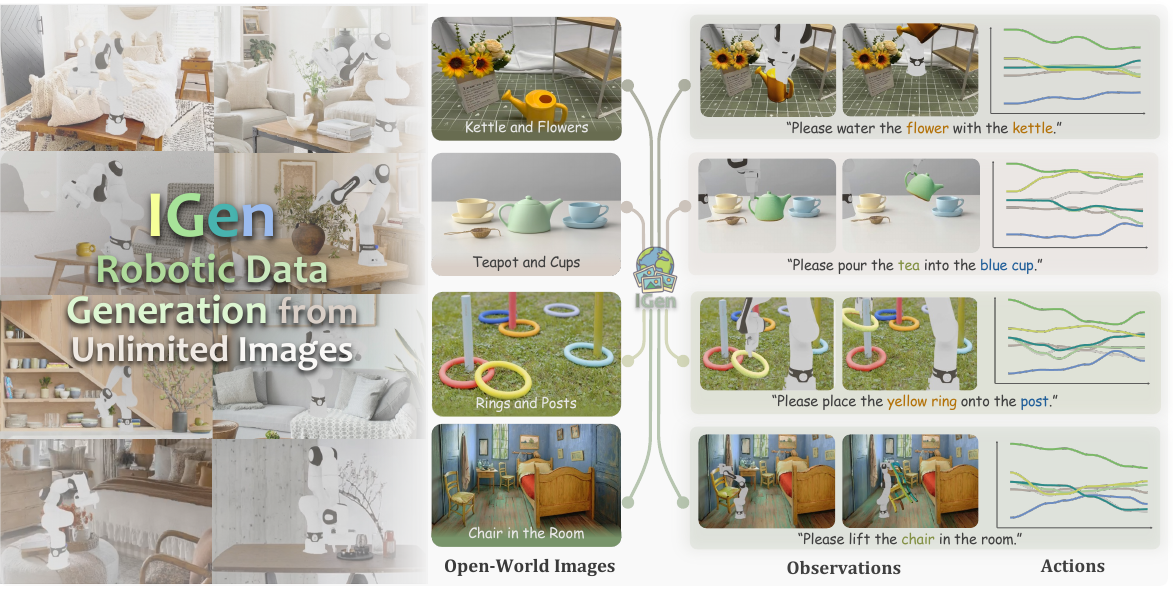}
    % \captionof{figure}{We propose \textbf{IGen}, a data generation framework that converts open-world images into grounded visuomotor data, enabling scalable data synthesis for robot learning via three components: (a) \emph{3D Scene Representation from Open-World Images} (Sec.~\ref{sec:pix23d}), (b) \emph{Spatial Planning for Behavior Generation} (Sec.~\ref{sec:spa}), and (c) \emph{Experience Synthesis for Robot Learning} (Sec.~\ref{sec:exp}). 
    % }
    \vspace{-0.2cm}
    \captionof{figure}{We propose \textbf{IGen}, a data generation framework that converts open-world images into grounded visuomotor data, enabling scalable data synthesis for robot learning. From a single image, IGen generates large-scale realistic observations and reliable actions. The policies trained solely on IGen-generated data can effectively generalize to real-world scenes and successfully perform manipulation tasks.
    }
    \label{fig:teaser}
\end{center}
}]

\maketitle

\begingroup
\renewcommand\thefootnote{\fnsymbol{footnote}}
\footnotetext[1]{Equal contribution.} 
\footnotetext[2]{Corresponding author.}
\endgroup

\begin{abstract}

\vspace{-0.12cm}
The rise of generalist robotic policies has created an exponential demand for large-scale training data. However, on-robot data collection is labor-intensive and often limited to specific environments. In contrast, open-world images capture a vast diversity of real-world scenes that naturally align with robotic manipulation tasks, offering a promising avenue for low-cost, large-scale robot data acquisition. Despite this potential, the lack of associated robot actions hinders the practical use of open-world images for robot learning, leaving this rich visual resource largely unexploited. To bridge this gap, we propose \textbf{IGen}, a framework that scalably generates realistic visual observations and executable actions from open-world images. IGen first converts unstructured 2D pixels into structured 3D scene representations suitable for scene understanding and manipulation. It then leverages the reasoning capabilities of vision-language models to transform scene-specific task instructions into high-level plans and generate low-level actions as $SE(3)$ end-effector pose sequences.  From these poses, it synthesizes dynamic scene evolution and renders temporally coherent visual observations. Experiments validate the high quality of visuomotor data generated by IGen, and show that policies trained solely on IGen-synthesized data achieve performance comparable to those trained on real-world data. This highlights the potential of IGen to support scalable data generation from open-world images for generalist robotic policy training. 
\end{abstract}
\vspace{-0.1cm} 
\section{Introduction}
\label{sec:intro}

Visuomotor policy learning~\cite{chi2023diffusion,ze20243d,black2024pi_0,bjorck2025gr00t,kim2024openvla} has shown great promise in enabling robots to perform open-world manipulation, yet it typically requires large-scale paired visual-action data for effective learning. Unfortunately, collecting such data on real robots is labor-intensive and often limited to specific environments. The high cost and environment-specific nature of robotic data collection remain a fundamental bottleneck to the generalization of visuomotor policies across diverse real-world scenarios~\cite{mirchandani2024so}.

In contrast to on-robot data collection, open-world images can be acquired at extremely low cost and encompass a vast diversity of scenarios that naturally align with real-world robotic tasks. Harnessing such rich visual resources offers a promising path toward building scalable, general-purpose robotic policies~\cite{zhang2024vision, laurenccon2024matters}. Indeed, the remarkable success of large vision-language models (VLMs)~\cite{achiam2023gpt, bai2025qwen2,lu2024deepseek} has demonstrated that open-world images provide a powerful foundation for training capable, generalist perception systems. However, robotic policy learning imposes unique demands: it requires not only semantically meaningful visual inputs but also physically grounded, executable action sequences~\cite{chi2023diffusion,zhao2023learning,fu2024mobile}. This mismatch prevents the effective utilization of open-world images for robot learning.

To address the complementary limitations of open-world images and robot-collected data, a growing line of work seeks to derive robot-relevant action representations from unstructured visual data. However, existing approaches remain limited in their ability to harness in-the-wild images for scalable robot learning. Previous \textit{real-to-sim-to-real} methods~\cite{torne2024robot,li2024robogsim,qureshi2025splatsim,lou2025robo,jia2025discoverse} require explicit reconstruction of physical workspaces to build simulation environments for data generation. This reliance on scene-specific acquisition prevents them from leveraging arbitrary open-world images, fundamentally limiting their scalability. Meanwhile, recent works~\cite{bruce2024genie, yang2023learning, zhou2024robodreamer, zhang2024combo, zhen2025tesseract, jang2025dreamgen} leverage vision generative models to predict future visual observations and robot actions within real-world scenes. However, due to the inherent limitations of video generation models, they cannot provide explicit robot actions and struggle to generate long-horizon or complex instruction-driven tasks.

In this work, we introduce \textbf{IGen}, a novel framework that takes robotic data synthesis the extra mile by enabling automated, grounded, and scalable visuomotor data generation from open-world images. It takes in-the-wild images as the sole visual input and leverages task and motion planning to automatically generate robot behaviors at scale without any human annotation. IGen further synthesizes large-scale visual observations and action trajectories that serve as high-quality training data for effective robot learning.

To be more specific, IGen employs the following unified pipeline to generate visuomotor data from open-world images. First, \duo{rather than directly operating on unstructured 2D pixels}, IGen leverages the strong visual capabilities of large vision models (LVM) to reconstruct scenes as 3D point clouds and spatial keypoints.  Next, it utilizes the visual understanding capabilities of vision-language models (VLM) to perform high-level task planning in 3D pixel space and transform motions into low-level control functions. 
Meanwhile, to generate robot-environment interactions, IGen uses the end-effector $SE(3)$ trajectory to perform rigid-motion-based synthesis of the scene point cloud sequence. Finally, frame-wise rendering is applied to the entire point cloud sequence, generating action-consistent visual observations of the target manipulation task.

We evaluate IGen along three dimensions: (1) visual fidelity, measured by the consistency between the generated visual observations and real-world perception;
(2) action quality, quantified by the scores of instruction following and physics alignment of the generated action videos; and (3) policy transferability, assessed through the performance of visuomotor policies trained on generated data and evaluated on real-world tasks. \duo{Experiments reveal that IGen generates visually realistic images and physically reliable robotic behaviors that align with specific task instructions, providing high-quality data for effective robot learning. Notably, results demonstrate that policies trained solely on IGen-generated data, without any human collection, can outperform those trained on real-world trajectories. 
}

In summary, our contributions are as follows:

\begin{itemize}
    \item We propose an effective data generation framework that produces scalable visual-action datasets from open-world images, integrating cross-scene generalization, instruction diversity and long-horizon task applicability.

    \item We transform unstructured open-world images into actionable 3D scene representations that enable robotic task reasoning and motion planning, generating task-consistent behaviors aligned with the physical world. Furthermore, we introduce a simulation-free point cloud synthesis approach that produces realistic visual observations and supports large-scale robot experience generation.

    \item We demonstrate that robotic policies trained on IGen-generated data can successfully perform real-world manipulation tasks without any additional data collection. \duo{This suggests the promising potential  of IGen to establish real-world images as an effective source for robot policy training, paving the way for scalable robot learning.}%, highlighting its potential to enable large-scale robot model training with minimal data acquisition cost, thereby lowering the barrier to scalable robot learning.

\end{itemize}

\section{Related Work}

\subsection{Visuomotor Robot Learning}

Recent progress in visuomotor policy learning ~\cite{chi2023diffusion,ze20243d,zhao2023learning,fu2024mobile, zheng2022extraneousness, hansen2022pre} enables robots to perform real-world manipulation tasks through end-to-end imitation learning from paired visual and action demonstrations. Inspired by the development of large-scale vision–language models (VLM)~\cite{achiam2023gpt, bai2025qwen2, lu2024deepseek}, recent efforts have focused on scaling up model size and data volume to build more generalist robotic models~\cite{team2024octo,huang2025otter,black2024pi_0,intelligence2025pi_,bjorck2025gr00t,kim2024openvla,team2025gemini, liu2025hybridvla, chen2025fast}. Meanwhile, many studies~\cite{huang2023voxposer, huang2024rekep, pan2025omnimanip} leverage the planning capabilities of VLMs to enable open-world robotic manipulation. Existing datasets~\cite{teed2021droid, o2024open, ebert2021bridge, fang2023rh20t} have collected large-scale demonstration data across diverse real-world scenes and task settings. Meanwhile, recent works ~\cite{chi2024universal, zeng2025activeumi, cheng2024open, ding2024bunny, gao2024efficient, mandlekar2023mimicgen, jiang2025dexmimicgen, xue2025demogen, zhang2025doglove} focus on improving the efficiency and scalability of robotic data collection. 

In contrast to vision–language models that benefit from abundant web-scale data, robotic learning lacks equivalent sources of large-scale, open-world demonstrations, posing a major bottleneck for developing generalist robot models~\cite{mirchandani2024so}. Open-world internet images provide a vast and diverse source of scenes for improving the generalization of robot learning. The core objective of our work is to enable scalable generation of robot-relevant data from such unstructured visual sources. 

\begin{figure*}
    \centering
    \includegraphics[width=\linewidth]{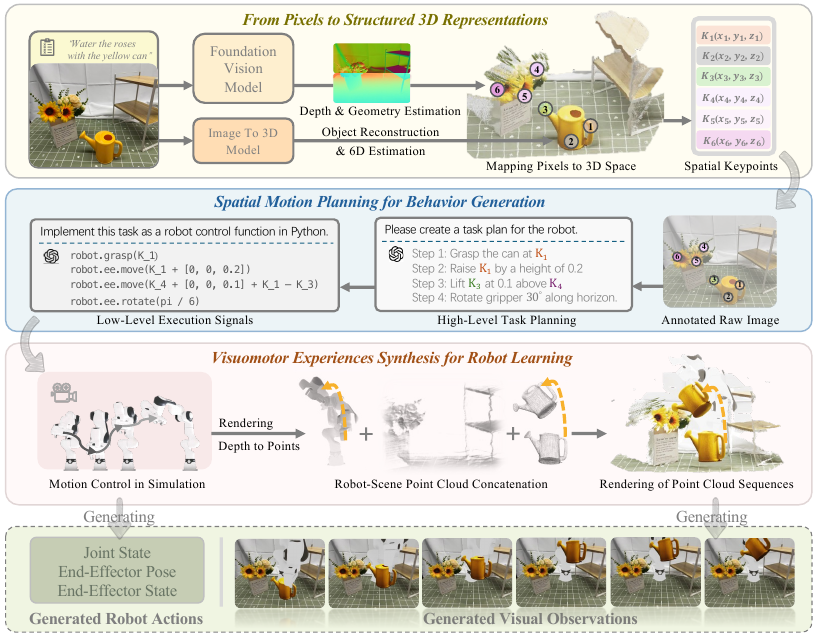}
    \caption{\textbf{Overview of IGen.}  Given an open-world image and a task description, IGen first reconstructs the environment and objects as point clouds via Foundation Vision Models. After spatial keypoint extraction, VLM maps the task description to high-level plans and low-level control commands. During the robot’s execution in simulation, a virtual depth camera captures the motion point cloud sequences. The resulting end-effector pose trajectory is used to synthesize dynamic point-cloud sequences, which are then rendered frame-by-frame into visual observations of the manipulation. The final output consists of the generated robot actions and the visual observations. 
    }
    \label{fig:method_overview}
\end{figure*}

\subsection{Real-Sim-Real Data Generation}

The Real-Sim-Real approach leverages real-world visual observations to build simulation workspaces for scalable policy learning and real-world transfer. Many recent works~\cite{dai2024automated, zook2025grs, ye2025video2policy, patel2025real, jiang2022ditto, wang2023gensim, li2024evaluating} employ digital twin or digital cousin approaches to recreate objects and spatial layouts for building simulation environments for robot training. However, such methods often struggle to capture fine-grained scene details for realistic world modeling. Meanwhile, some approaches~\cite{torne2024robot,duan2023ar2,li2024robogsim,qureshi2025splatsim,lou2025robo,jia2025discoverse} leverage high-fidelity 3D reconstruction techniques to model real-world scenes from multi-view observations, which are instantiated as executable robot workspaces within simulation environments. These methods remain confined to task-specific environments and require extensive visual data collection for scene reconstruction. 

Open-source internet images provide a rich and scalable data source that can cover most real-world scenarios. To leverage this advantage, RoLA~\cite{zhao2025robot} proposes an open-world, image-based framework for robot learning data generation. However, it relies on accurate physical property estimation for scene modeling and remains limited in modeling complex interactive manipulation tasks. IGen, on the other hand, leverages images as the primary data source without explicit physics-based reconstruction, offering a lightweight alternative to simulation-based pipelines. 

\subsection{Experience Synthesis from Unstructured Data}

Inspired by the success of large-scale visual models, there is growing interest in leveraging vast amounts of unstructured internet data—such as images and videos—to provide scalable training sources for robot learning.  Recent approaches~\cite{singh2025hand, chen2024object, chen2025vidbot, lepert2025phantom, yuan2025hermes, zhu2024densematcher} synthesize robot manipulation data from collections of human demonstration videos from the Internet or real-world recordings. However, such data are limited in scene diversity and inherently suffer from human bias. Many works~\cite{yang2023learning, ko2023learning, zhen2025tesseract, zhou2024robodreamer, jang2025dreamgen} build upon large-scale pre-trained video generation models to create visual observations for robotic manipulation. Yet, these generative methods often lack grounded robotic actions, struggle with complex and long-horizon tasks, and incur substantial computational costs. Images are the most abundant source of data on the internet, yet they inherently lack robot-relevant information. To this end, IGen aims to transform unstructured image data into grounded robotic experiences, providing a scalable source of data for robot learning.

\section{Methodology}
\label{sec:method}

IGen is a robotic data generation framework that synthesizes task-specific robot actions and visual observations from open-world images. The architecture consists of three main stages: (1) \textbf{Scene Reconstruction}, converting the input image into a manipulable workspace for the robot; (2) \textbf{Action Planning}, reasoning over spatial keypoints to generate action trajectories; and (3) \textbf{Observation Synthesis}, composing and rendering point-cloud sequences to generate visual observations of the task. 

\subsection{From Pixels to Structured 3D Representations}
\label{sec:pix23d}

Open-world images are unstructured and lack explicit robot-related actions. Hence, the key to enabling robotic learning from images lies in transforming raw pixels into structured representations that robots can effectively interpret and take actions upon. To this end, we adopt 3D point clouds as the modality, which offers a format better suited to visual observation and structured editing.

To begin with, we employ a versatile monocular geometric foundation model~\cite{hu2024metric3d} to estimate the depth and geometric structure of the scene. We first use a VLM to analyze the prompt and identify task-relevant objects, and then employ Segment Anything Model \cite{kirillov2023segment, liu2024grounding} to obtain masks $\mathcal{M}$ of task-relevant objects. Following the keypoints representation in \cite{huang2024rekep}, we extract scene features using DINOv2~\cite{oquab2023dinov2}. 
Next, we apply K-means clustering to both the feature embeddings and 3D coordinates, resulting in a set of $K$ spatial keypoints, denoted as $\mathcal{K} = \{\, k_j\in \mathbb{R}^3\ \mid j = 1, 2, \dots, K \}$, with their associated 3D coordinates.

Then, the inpainting model~\cite{yu2023inpaint} is applied to the original image to remove the target manipulable object and reconstruct the background image $\mathbf{I}_{\mathrm{bg}}
 \in \mathbb{R}^{H \times W \times 3}$. Given the estimated depth map $\mathbf{D} \in \mathbb{R}^{H \times W}$ and the camera intrinsic matrix $\mathbf{K}$, all pixels are lifted into 3D space as a dense 3D point cloud $P_{\mathrm{bg}} \in \mathbb{R}^{H \times W \times 6}$. To address incomplete modeling of manipulated objects from monocular views, we employ a 3D generative reconstruction model~\cite{xiang2025structured} to their full 3D shape and appearance. After reconstructing the objects into dense point clouds, we re-position the completed objects to their original pose using 6D pose estimation~\cite{lee2025any6d}. Furthermore, to enable spatial domain randomization, we extract a set of feasible placement points from the supporting surface (e.g., the spatial points of the tabletop), which serve as candidate spatial poses for object generalization. Details of the reconstruction are provided in the appendix. 

\begin{figure*}
    \centering
    \includegraphics[width=\linewidth]{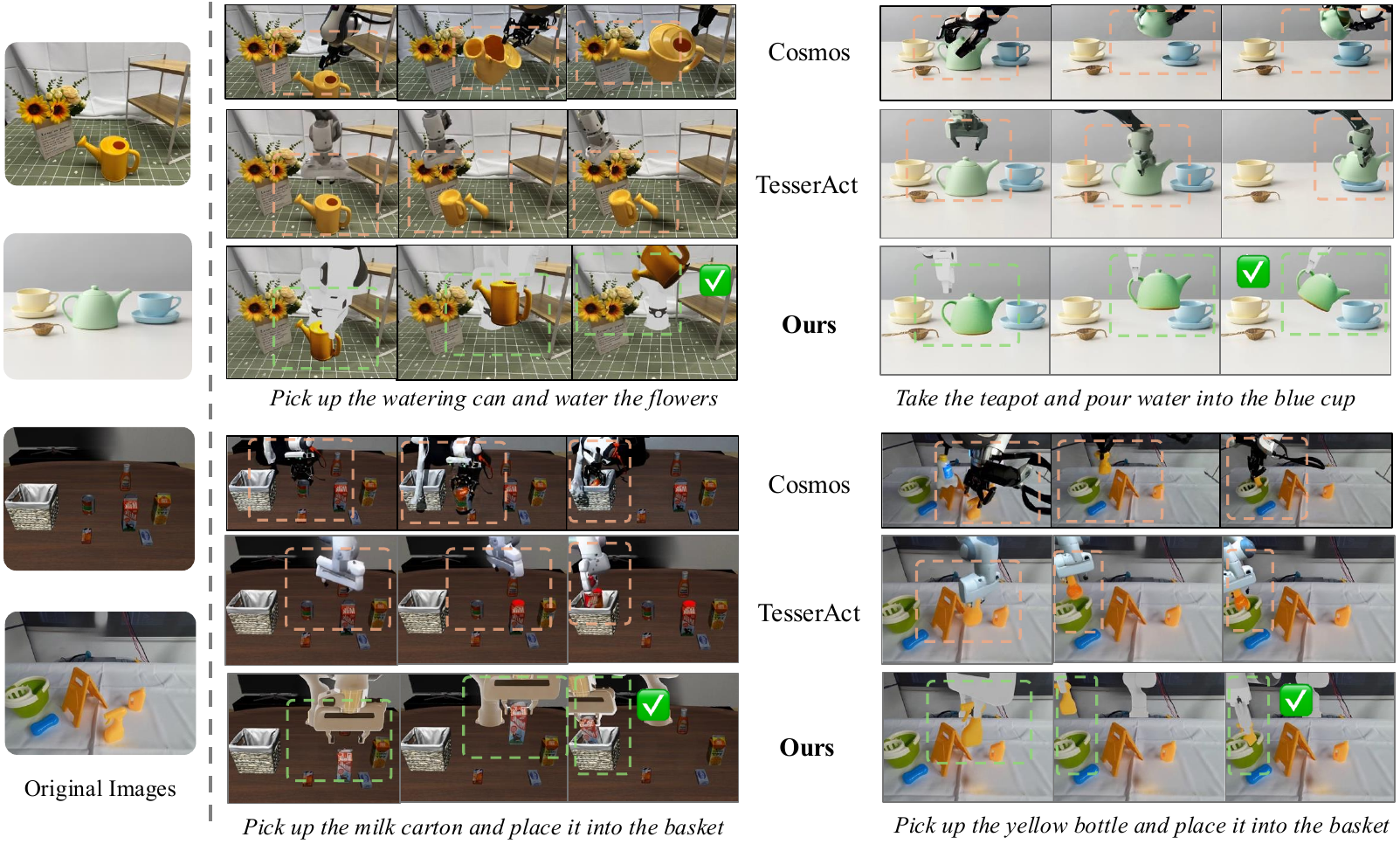}
    \caption{\textbf{Qualitative comparison of robotic behavior generation using IGen.} Given a single captured image and a natural-language manipulation instruction, TesserAct~\cite{zhen2025tesseract}, Cosmos~\cite{agarwal2025cosmos}, and our IGen generate behavior observations. IGen produces more instruction-consistent and physically coherent object motions, closely matching the intended tasks. The green box represents action observations that adhere to physical laws and follow the task instructions, and the checkmark indicates task completion. 
    }
    \label{fig:fig_3}
\end{figure*}

\subsection{Spatial Planning for Behavior Generation}
\label{sec:spa}
Since open-world images lack action-related guidance for robots, we leverage the strong visual reasoning and planning capabilities of VLMs~\cite{achiam2023gpt, bai2025qwen2} to guide robotic behavior generation. We provide the image annotated with keypoints $\mathcal{K}$, their 3D coordinates, and the task instruction as inputs to the VLM. The VLM decomposes the overall task into a set of sub-stages $\mathcal{S} = \{S_i\}_{i=1}^N$ through high-level task planning. For each sub-stage, the VLM model generates an action description associated with the keypoints.

To enable robotic action execution, we develop an easily programmable control language in Python based on the end-effector’s $SE(3)$ pose, translating high-level task stages $\mathcal{S}$ into executable low-level control functions $\mathcal{F}=\{f_i\}_{i=1}^N$ with VLM. For each task stage $S_i$, the function $f_{i}$ is defined as a keypoint-conditioned solver that computes the reference end-effector pose from the spatial anchors in $\mathcal{K}_i$. Here, function parameters such as translation distance and lift height are inferred by the VLM from 3D keypoint coordinates.
During the pre-manipulation stage, the end-effector interacts precisely with the object based on a grasping prior model, where the end-effector pose is predicted by the grasp model~\cite{fang2020graspnet, murali2025graspgen}, and the gripper state is determined by the point cloud width along the gripper’s principal axis. During manipulation, keypoints on the manipulated object are treated as movable points rigidly attached to the end-effector, while other keypoints serve as static scene anchors. The specific prompts used are detailed in the appendix.

Starting from the initial robot state $x_0$, the motion planner produces a set of sub-goals $\mathcal{X} = \{\hat{x}_i\}_{i=1}^N$ that define the desired states for subsequent control execution. Each stage computes the next state as $\hat{x}_{i} = f_{i}(\hat{x}_{i-1})$, where $\hat{x}_0 = x_0$. We use a motion planner to generate feasible trajectories, which are then executed in a simulation environment, producing an action sequence $\mathcal{A} =\{ {a_t}\}_{t=1}^T$ over $T$ time steps, sampled at a fixed frame rate. Each action $a_t$ includes the robot's end-effector pose and joint positions. Detailed simulation settings are provided in the appendix.

\begin{table*}[t]
\centering
\resizebox{\textwidth}{!}{
\begin{tabular}{@{}l c c c c c c c c c c@{}}
\toprule
\multirow{2}{*}{\textbf{Task}} & 
\multicolumn{2}{c}{\textbf{PSNR}$\uparrow$} & 
\multicolumn{2}{c}{\textbf{SSIM}$\uparrow$} & 
\multicolumn{2}{c}{\textbf{LPIPS\textsubscript{1}}$\downarrow$} & 
\multicolumn{2}{c}{\textbf{LPIPS\textsubscript{2}}$\downarrow$} &
\multicolumn{2}{c}{\textbf{LPIPS\textsubscript{3}}$\downarrow$} \\
\cmidrule(lr){2-11}
& Real-to-Sim & \textbf{IGen} 
& Real-to-Sim & \textbf{IGen} 
& Real-to-Sim & \textbf{IGen}
& Real-to-Sim & \textbf{IGen}
& Real-to-Sim & \textbf{IGen} \\
\midrule
Pick Carrot on Plate 
& 18.1480 & \cellcolor[HTML]{F2F7F2}\textbf{28.2611}
& 0.6756 & \cellcolor[HTML]{F2F7F2}\textbf{0.8371}
& 0.2864 & \cellcolor[HTML]{F2F7F2}\textbf{0.0518}
& 0.3882 & \cellcolor[HTML]{F2F7F2}\textbf{0.1166}
& 0.2057 & \cellcolor[HTML]{F2F7F2}\textbf{0.0418} \\
Put Eggplant in Basket 
& 16.6910 & \cellcolor[HTML]{F2F7F2}\textbf{23.2821}
& 0.7706 & \cellcolor[HTML]{F2F7F2}\textbf{0.8350}
& 0.2458 & \cellcolor[HTML]{F2F7F2}\textbf{0.0825}
& 0.2658 & \cellcolor[HTML]{F2F7F2}\textbf{0.1242}
& 0.1682 & \cellcolor[HTML]{F2F7F2}\textbf{0.0547} \\
Put Spoon on Towel 
& 15.9262 & \cellcolor[HTML]{F2F7F2}\textbf{26.7524}
& 0.6041 & \cellcolor[HTML]{F2F7F2}\textbf{0.8621}
& 0.4275 & \cellcolor[HTML]{F2F7F2}\textbf{0.0621}
& 0.4879 & \cellcolor[HTML]{F2F7F2}\textbf{0.1129}
& 0.2980 & \cellcolor[HTML]{F2F7F2}\textbf{0.0433} \\
Stack Cubes 
& 18.2944 & \cellcolor[HTML]{F2F7F2}\textbf{29.7206}
& 0.6828 & \cellcolor[HTML]{F2F7F2}\textbf{0.8747}
& 0.3345 & \cellcolor[HTML]{F2F7F2}\textbf{0.0558}
& 0.4419 & \cellcolor[HTML]{F2F7F2}\textbf{0.1021}
& 0.2491 & \cellcolor[HTML]{F2F7F2}\textbf{0.0386} \\
\midrule
\textbf{Average}
& 17.2649 & \cellcolor[HTML]{E2ECE2}\textbf{27.0040}
& 0.6833 & \cellcolor[HTML]{E2ECE2}\textbf{0.8522}
& 0.3235 & \cellcolor[HTML]{E2ECE2}\textbf{0.0630}
& 0.3959 & \cellcolor[HTML]{E2ECE2}\textbf{0.1139}
& 0.2302 & \cellcolor[HTML]{E2ECE2}\textbf{0.0446} \\
\bottomrule
\end{tabular}}
\caption{We compare the visual similarity between the digital-twin scenes reconstructured by Real-to-Sim and those generated by IGen. %Simpler and those generated by IGen, against the real-world scenes. 
Real-to-Sim refers to the method in Simpler~\cite{li2024evaluating} that converts real-world scenes into simulated digital-twin scenes. %their simulation counterparts. 
We compute the LPIPS~\cite{wigner1990unreasonable} variants using AlexNet~\cite{krizhevsky2012imagenet}, VGGNet~\cite{simonyan2014very}, and SqueezeNet~\cite{iandola2016squeezenet}, denoted as LPIPS\textsubscript{1}, LPIPS\textsubscript{2}, and LPIPS\textsubscript{3}.
$\uparrow$\,/\,$\downarrow$ indicates higher\,/\,lower is better.}
\label{tab:scene_results}
\end{table*}

\begin{figure*}[t]
    \centering

    % --- 左侧大图组 ---
    \begin{minipage}[t]{0.58\textwidth}

        \centering
        \includegraphics[width=\textwidth]{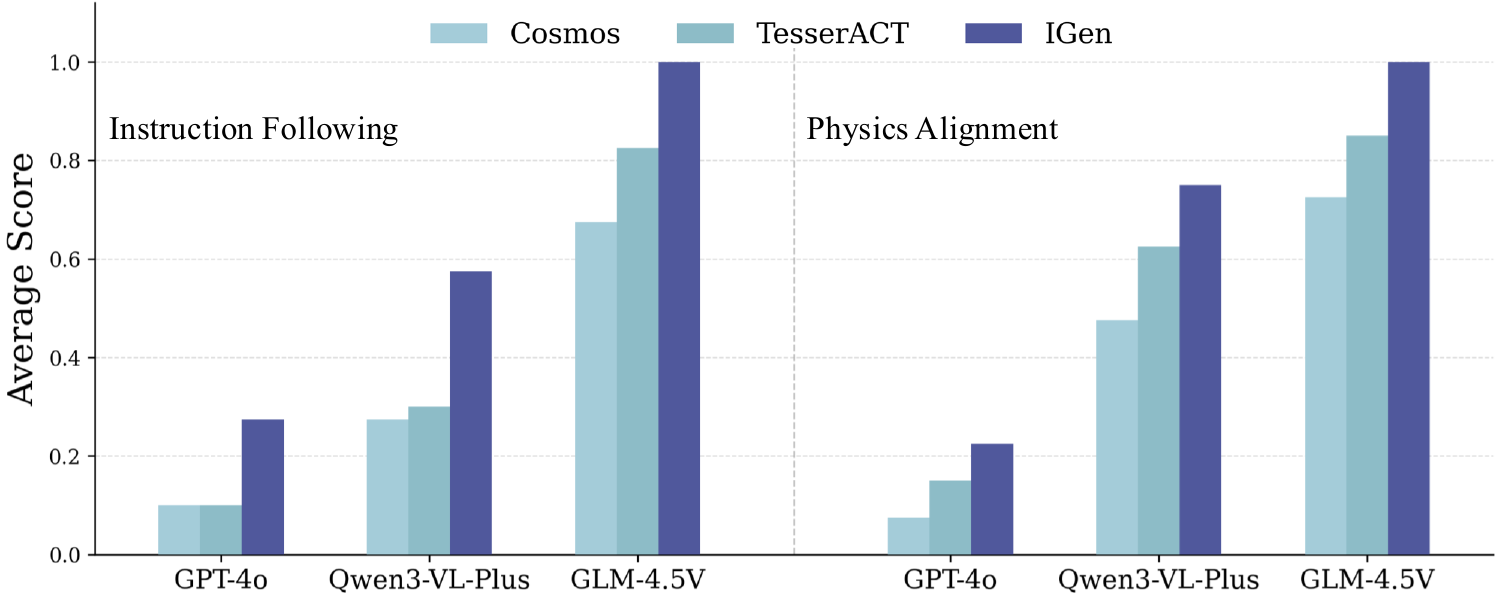}
        \label{subfig:ablation_mean_acc}
        \hfill

        \vspace{-0.3cm}
        \caption{
            \textbf{Quantitative Comparison of robotic behavior generated by IGen.} Performance is assessed on DreamGen Bench~\cite{jang2025dreamgen} under two criteria: \textit{Instruction Following} and \textit{Physics Alignment}. Evaluations are conducted using GPT-4o~\cite{achiam2023gpt}, Qwen-3-VL-Plus~\cite{bai2025qwen2} and GLM-4.5V~\cite{zeng2025glm} as video assessment models. Each method generates 40 videos along with the prompts, and the reported metric represents the proportion of videos receiving a score of 1 from the evaluator. 
        }
        \label{fig:fig_4}
    \end{minipage}
    \hfill
    % --- 右侧单图 ---
    \begin{minipage}[t]{0.41\textwidth}
        \centering
        \includegraphics[width=\textwidth]{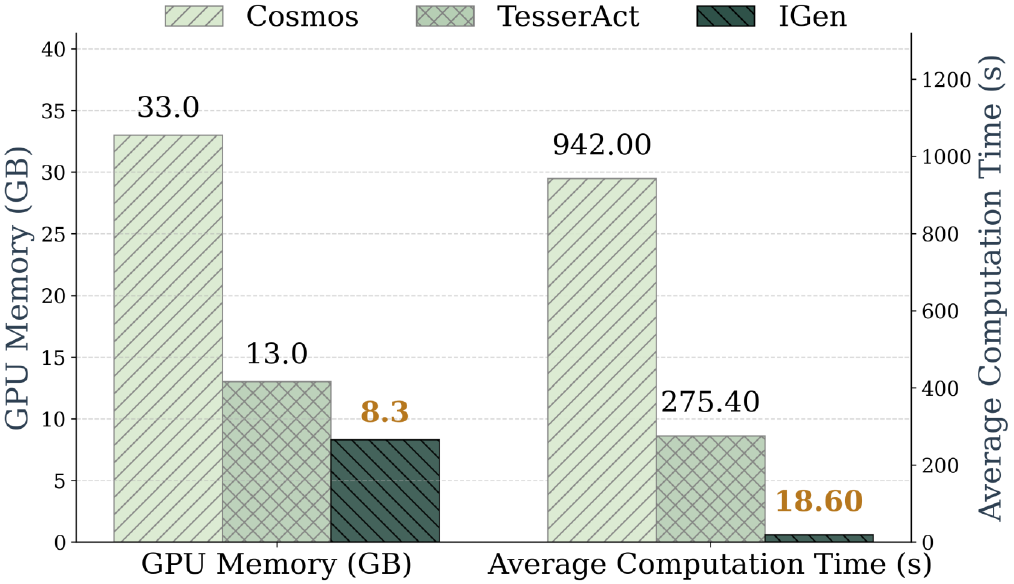}
        \vspace{-0.3cm}
        \caption{
            \textbf{Evaluation of IGen’s computational efficiency.} We compare the video generation time and GPU memory consumption of IGen and baselines under identical input images and task instructions. The average computation time refers to the time required to generate one robot behavior video.
        }
        \label{fig:fig_5}
    \end{minipage}

\vspace{-0.3cm}

\end{figure*}

\subsection{Experience Synthesis for Robot Learning}
\label{sec:exp}

To obtain visual observations synchronized with embodied actions, 
we propose a robotic experience synthesis framework based on real-time point cloud rendering. 
A complete point cloud observation of a manipulation task comprises the action point cloud sequence of the robot 
and the dynamic point cloud sequence of the scene. 

To synthesize action point clouds, the simulated robot is placed at the planned pose 
\(p_{\text{robot}}\) in simulation, 
and a virtual camera is positioned at \(p_{\text{cam}}\), 
aligned with the viewpoint used for generating the scene point clouds. From the action sequence $\mathcal{A}$, we obtain an end-effector pose trajectory
$\mathcal{T} = \{\mathbf{T}_{t}\}_{t=1}^T$, where $\mathbf{T}_{t} \in SE(3)$ denotes the 6-DoF pose of the end-effector at time step $t$, 
including both rotation and translation. 
At each time step $t$, the environment is rendered to produce synchronized RGB and depth frames, which are then back-projected through the virtual camera $\mathbf{C}$ to construct the point cloud sequence of the robot's motion \(
\mathcal{P}_{\text{robot}} = \{P_{\text{robot},t}\}_{t=1}^{T}\). Meanwhile, the background is modeled as a static point cloud sequence \( \mathcal{P}_{\text{bg}}\). 

We perform dynamic interaction between the end-effector and the point clouds based on transformations of the end-effector's pose. Assume the grasp is established at time \(t_g\), with the object's pose
\(\mathbf{T}_{\text{obj}, t_g}\) and the end-effector pose 
\(\mathbf{T}_{t_g}\). For all time steps \(t \in \mathcal{T}_{\text{grasp}}\), where  \(\mathcal{T}_{\text{grasp}}\) denotes the set of time indices during which the gripper remains closed on the object, 
the world pose the object evolves by rigidly following the end-effector. The manipulated object, represented as a point cloud sequence \( \mathcal{P}_{\text{obj}} = \{P_{\text{obj}, t}\}_{t=1}^{T}\), undergoes rigid-body transformations induced by the end-effector poses. The transformation at time $t$ can be expressed as:
\begin{equation}
P_{\text{obj}, t} =
\begin{cases}
P_{\text{obj}, t} & t \notin \mathcal{T}_{\text{grasp}}, \\[6pt]
\mathbf{T}_{t}\,(\mathbf{T}_{t_g})^{-1}\,\mathbf{T}_{\text{obj}, t_g}\,P_{\text{obj}, t_g} & t \in \mathcal{T}_{\text{grasp}}~.
\end{cases}
\end{equation}

By combining the static environment, the robot, and the manipulated object, we represent the complete task as a composite point cloud sequence:
\begin{equation}
\mathcal{P}_{\text{task}} = \mathcal{P}_{\text{bg}} \cup \mathcal{P}_{\text{obj}} \cup \mathcal{P}_{\text{robot}}.
\end{equation}

By rendering the point cloud sequence through the virtual camera $\mathbf{C}$, 
we obtain the visual observations $\mathcal{O}$. 
The temporally synchronized observations, together with the action sequence $\mathcal{A}$, 
constitute paired visual--action data for robot learning.

\section{Experiment}

In this section, we aim to address the following three research questions:
(1) Can \textbf{IGen} generate visually realistic data from real-world images across various scenes?
(2) Can \textbf{IGen} efficiently generate robot actions that align with the environment and follow task instructions?
(3) Can \textbf{IGen} synthesize effective robotic training data directly from a single image, enabling policy training and real-world deployment without any human-teleoperated demonstrations?

\begin{figure*}
    \centering
    \vspace{-0.5cm}
    \includegraphics[width=\linewidth]{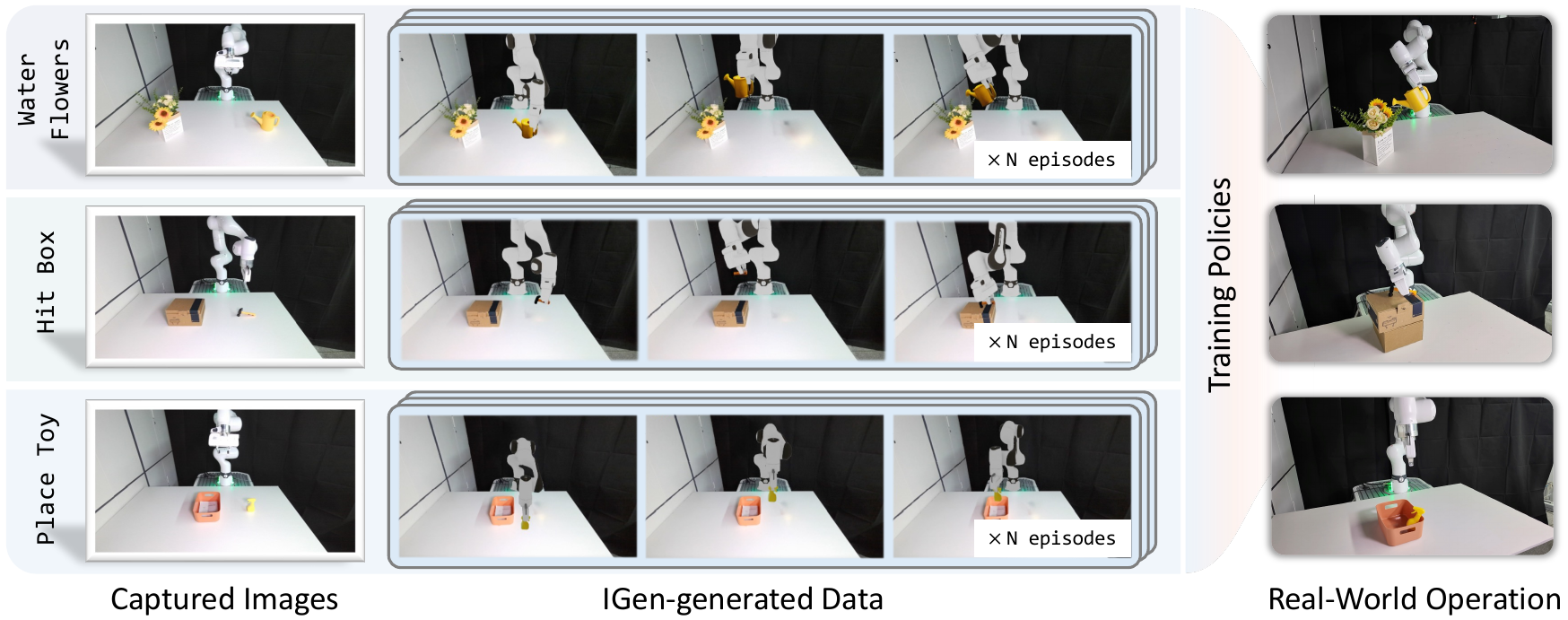}
    \vspace{-0.5cm}
    \caption{\textbf{Real-world Experiments.} Starting from a captured real-world scene image, IGen automatically generates 1,000 task demonstrations with spatial randomization. The resulting data are used to train a visuomotor policy, which is later deployed and evaluated in the real world. We evaluate our method on real-world tasks including ``Water Flowers'', ``Hit Box'' and ``Place Toy''.
    }
    \vspace{-0.2cm}
    \label{fig:fig_6}
\end{figure*}

\subsection{Scene Reconstruction Fidelity}

In this section, we compare IGen with Real-to-Sim methods to measure the visual realism of the synthesized scenes. We conduct experiments on the Simpler dataset~\cite{li2024evaluating}, which provides diverse real-world robotic manipulation scenes. Based on the real-world captured images in Simpler, we employ IGen to reconstruct the corresponding 3D scenes for evaluation. Meanwhile, the Real-to-Sim method in Simpler reconstructs digital-twin scenes from real-world images. 

We adopt multiple visual metrics, including PSNR, SSIM, and LPIPS~\cite{wigner1990unreasonable}, to assess the visual consistency between the original real-world scenes and those reconstructed by the Real-to-Sim method in Simpler or IGen. As shown in Tab.~\ref{tab:scene_results}, IGen achieves superior performance across multiple evaluation metrics. For instance, using LPIPS scores computed across multiple perception models to assess the perceptual discrepancy between generated scenes and the original images, we observe that IGen achieves up to a \textbf{5.13$\times$} improvement in similarity compared with the Real-to-Sim baseline. These results indicate that the scenes generated by IGen remain highly consistent with the corresponding real-world images, demonstrating its capability to produce visually faithful and realistic observations that align closely with real environments, thus mitigating the sim-to-real gap.

\subsection{Evaluation of Behavior Generation}

This section focuses on evaluating the behavioral accuracy and instruction alignment of IGen’s generated robotic behaviors. We evaluate IGen across diverse visual sources and task instructions to assess the quality of robotic behaviors generated from a single input image. For comparison, we include Cosmos-Predict2~\cite{agarwal2025cosmos, jang2025dreamgen} and TesserAct~\cite{zhen2025tesseract} as baselines, both of which are capable of generating robot behaviors from a single image input. Cosmos-Predict2 employs the Cosmos-Predict2-2B-Video2World model, while TesserAct uses the model fine-tuned from CogVideoX-5B-I2V~\cite{yang2024cogvideox}. 
For open-world images with unknown intrinsics, we adopt a canonical camera and use the open-world intrinsics provided by Metric3Dv2~\cite{hu2024metric3d}.

\begin{figure}
    \centering
    \includegraphics[width=\linewidth]{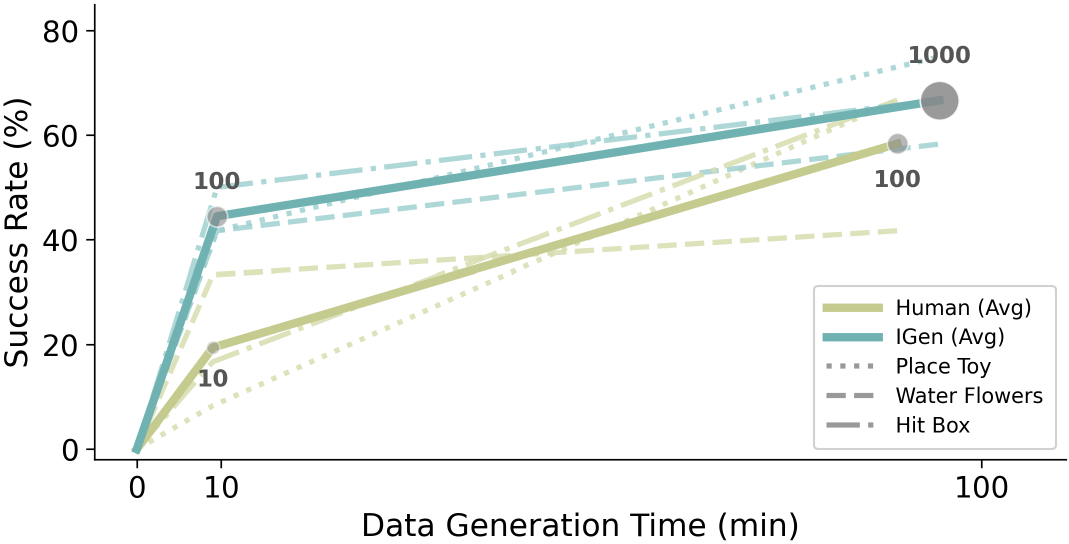}
    \vspace{-0.5cm}
    \caption{\textbf{Real-world robot evaluation results.} Policies are evaluated under five settings: zero-shot, 10 human-teleoperated samples, 100 human-teleoperated samples, 100 IGen-generated samples, and 1,000 IGen-generated samples. The figure reports both per-task performance (thin lines) and the average across all tasks (thick lines). Compared with human teleoperation, IGen-synthesized data can generate substantially more data within a similar time budget and achieve higher success rates.
    }
    \vspace{-0.5cm}
    \label{fig:fig_7}
\end{figure}

\vspace{1mm}\noindent\textbf{Qualitative Experiments.} As shown in Fig.~\ref{fig:fig_3}, we evaluate on unlabeled, open-world images without any task annotations or demonstrations. Each image is associated with a scene-specific instruction, which serves as a high-level command for IGen and as the language prompt for Cosmos-Predict2 and TesserAct. The results demonstrate that IGen can generate complete and coherent robotic behaviors that accurately follow the given task instructions, producing physically consistent motions and visual observations that align with real-world physics. In contrast, the baseline methods exhibit temporal discontinuities, geometric distortions, and inaccurate motion execution, failing to produce instruction-aligned behaviors. 

\noindent\textbf{Quantitative Experiments.}~To comprehensively evaluate the quality of robotic behaviors generated by IGen, we conduct experiments on the DreamGen Bench~\cite{jang2025dreamgen}. Two criteria---\textit{Instruction Following} and \textit{Physics Alignment}---are used to assess the quality of the generated robot-manipulation videos. We employ GPT-4o~\cite{achiam2023gpt}, Qwen-3-VL-Plus~\cite{bai2025qwen2}, and GLM-4.5V~\cite{zeng2025glm} as evaluation models, ensuring fair and consistent comparison under identical video resolution and frame-rate settings. As shown in Fig.~\ref{fig:fig_4}, IGen outperforms all compared models in both \textit{Instruction Following} and \textit{Physics Alignment}. As shown in the results, IGen produces \textbf{nearly twice} as many \textit{Instruction Following} successes as the baseline when evaluated by Qwen-3-VL-Plus. 
% Furthermore, under the \textit{Physics Alignment} metric, GLM-4.5V judges \textbf{100\%} of IGen-generated videos to exhibit physically consistent dynamics.
These results indicate that IGen provides high-quality visual observations and physically accurate action sequences, which can effectively support robot learning in real-world environments.

\noindent\textbf{Computational Efficiency.}~For large-scale model training, generating data at scale requires high computational efficiency. In this section, we evaluate IGen’s computational efficiency, focusing on the runtime performance and resource consumption. For open-world image inputs, we use IGen to generate 1,000 randomized samples per image and report the average per-sample generation time and GPU memory utilization. For comparison, we compute the average per-sample video generation time and GPU memory utilization for Cosmos-Predict2 and TesserAct. As illustrated in Fig.~\ref{fig:fig_5}, IGen demonstrates higher efficiency in data generation, requiring only 8.3 GB of GPU memory and approximately 18.6 seconds per sample. Under the same GPU memory conditions, IGen achieves data generation with approximately \textbf{30$\times$} and \textbf{200$\times$} higher efficiency than TesserAct and Cosmos-Predict2, respectively. 
These results highlight that IGen is significantly more computationally efficient and  scalable for large-scale robotic data synthesis.

\subsection{Robot Learning from Unstructured Images}

We provide IGen with a single image and, without any on-robot data, automatically generate visuomotor data to train the robot model for real-world manipulation tasks. 

\vspace{1mm}\noindent\textbf{Hardware Setup.}~All real-world experiments are conducted on a Franka Research 3 robot arm. Perception is provided by a single Microsoft Kinect RGB-D camera mounted in front of the manipulator, from which we use only the RGB stream as visual input.

\vspace{1mm}\noindent\textbf{Robot Policy.}~We adopt $\pi_0$~\cite{black2024pi_0} as the base vision-language-action model. We first verify that, without any fine-tuning, $\pi_0$ achieves an almost zero success rate across our tasks, indicating limited zero-shot transfer capability in our experimental setting. Subsequently, the model is fine-tuned separately on teleoperation data and on IGen-generated data with LoRA~\cite{hu2022lora}, and evaluated on the same set of manipulation tasks for comparison. 

\vspace{1mm}\noindent\textbf{Tasks and Evaluation.} We design diverse manipulation tasks to evaluate the effectiveness of IGen, covering three types: Placement (\textit{“Place the bottle”}), Scene-Object Interaction (\textit{“Water the flowers”}), and Object-Object Interaction (\textit{“Hit Box”}). We assess the effect of data scaling by training the model with (a) 10 real-robot samples, (b) 100 real-robot samples, (c) 100 IGen-generated samples, and (d) 1,000 IGen-generated samples, and evaluating them under the same conditions.

\vspace{1mm}\noindent\textbf{Data Generation.}~Starting from a single camera-view image, IGen applies spatial randomization to generate training data for real-world scenes. A manipulable tabletop workspace is designated within the robot’s reachable area, and 3D points within this region are randomly sampled from the scene point cloud as object placement positions. As shown in Fig.~\ref{fig:fig_6}, we demonstrate large-scale automated data generation from a single image across multiple tasks, and train robot policies to execute real-world tasks.

\vspace{1mm}\noindent\textbf{Performance Analysis.}~As shown in Fig.~\ref{fig:fig_7}, we compare zero-shot performance (0 sample), teleoperation-based fine-tuning (10 and 100 samples), and IGen-based fine-tuning (100 and 1,000 samples). When directly deploying the $\pi_0$ model without any fine-tuning samples, the success rate remains close to zero. With just a single image, IGen automatically generates synthetic data with strong generalization capability for policy fine-tuning. On \textit{Place Bottle} task, the fine-tuned model achieves a significant increase in task success rate, increasing \textbf{from 0.0\% to 75.0\%}. 
Notably, with 100 and 1,000 IGen-generated samples, the policy, \textbf{without any human-collected data fine-tuning}, achieves \textbf{success rates of 44.5\% and 66.7\%}, respectively, surpassing the 19.4\% and 58.3\% achieved by human teleoperation under the same data collection time budget. 
We find that the policy fine-tuned on a large scale of IGen-generated data demonstrates robust performance in executing task trajectories. 
% The result suggest that IGen can serve as an effective and scalable alternative to human teleoperation for training robot policies. 
These findings highlight the potential to enable robots to learn from a wide range of open-world images, further enhancing its applicability for diverse real-world tasks.

\section{Conclusion}

In this work, we propose a scalable data generation framework that automatically transforms open-world images into high-quality visuomotor data for robot learning. Leveraging only unstructured open-world images, our approach enables the scalable generation of training data, eliminating the need for human effort while effectively improving robot policy performance. This work has the potential to alleviate the challenge of limited real-world robot data, offering a promising data-driven solution for the development of generalist robot policies.

\section*{Acknowledgements}
We thank the anonymous CVPR reviewers and ACs for their invaluable feedback. 
We would also like to thank Yuzhi Huang from Xiamen University, Yong Zhong from Southern University of Science and Technology, and Xin Xiang and Zhixing Zhang from South China University of Technology for their valuable suggestions and support.
This work was supported by National Natural Science Foundation of China (Grant No.~92467204 and 62472249), and Shenzhen Science and Technology Program (Grant~No. JCYJ20220818101014030 and KJZD20240903102300001).

{
    \small
    \bibliographystyle{ieeenat_fullname}
    \bibliography{main}
}

\clearpage
\setcounter{page}{1}
\maketitlesupplementary

\section{Single-Image Scene Reconstruction Details}

\begin{figure}[!htbp]
    \centering
    \includegraphics[width=0.99\linewidth]{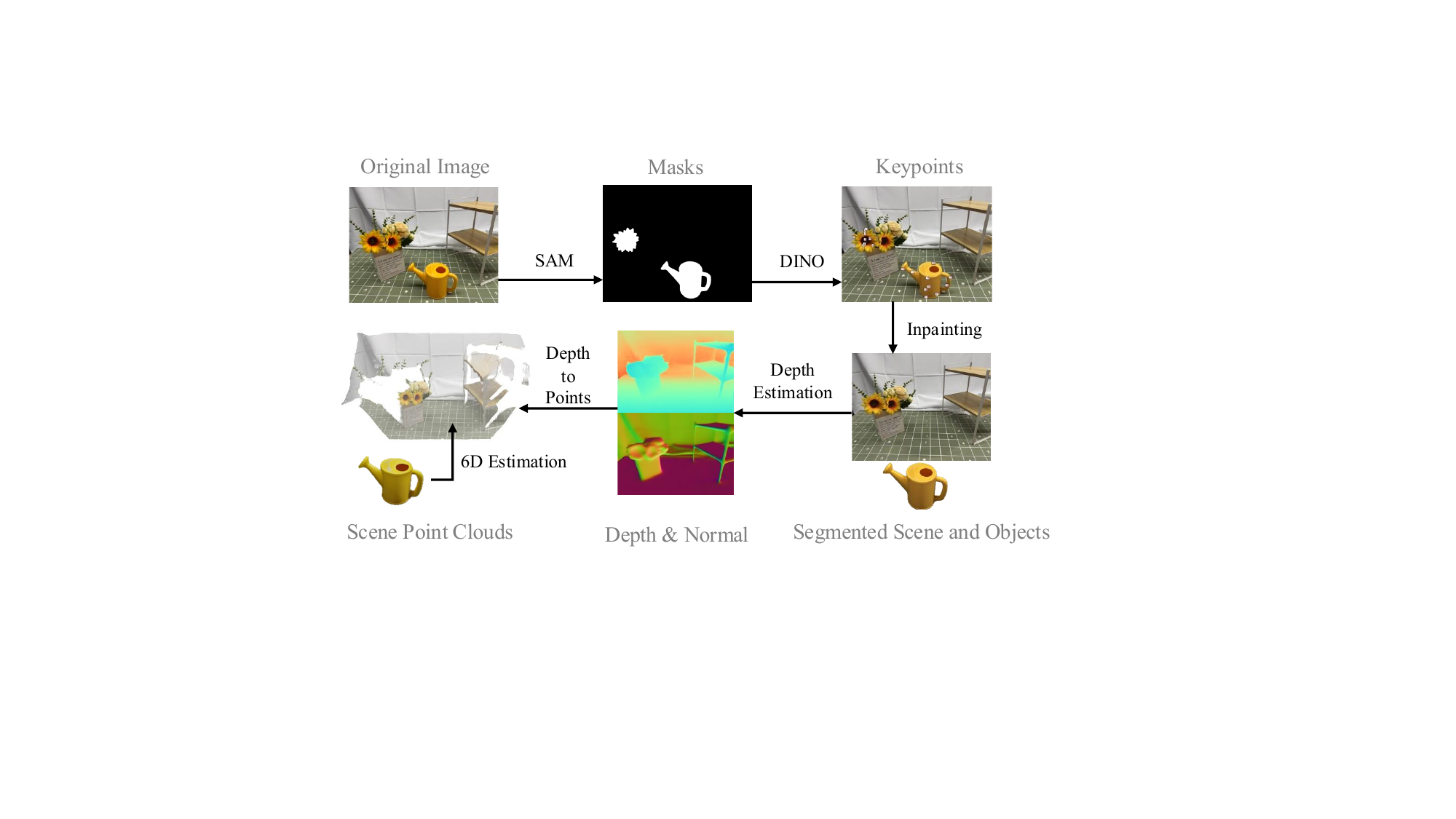}
    \caption{\textbf{Single-Image Scene Reconstruction Pipeline.} 
    }
    \label{fig:fig_sup_1}
\end{figure}

In this section, we describe how IGen reconstructs the 3D scene from a single open-world RGB image to facilitate robot data generation, as illustrated in Fig.~\ref{fig:fig_sup_1}.

We use Metric3Dv2~\cite{hu2024metric3d} to estimate the depth and convert image pixels into a point cloud. 
For open-world images, the focal length is fixed to $1000$, while for specific camera types (e.g., iPhone or Microsoft Kinect Camera), we adopt their corresponding intrinsic parameters. 
The resulting point cloud preserves the same spatial resolution and dimensions as the original RGB image.

For object-level reconstruction, we utilize the TRELLIS model~\cite{xiang2025structured} to perform monocular 3D reconstruction and convert the outputs into colored point clouds. 
The input images to TRELLIS are pre-processed using segmentation masks obtained from SAM~\cite{kirillov2023segment}, ensuring that the reconstruction focuses on the target objects.

\vspace{-0.3cm}
\vspace{-0.3cm}

\section{Simulation Environment Details}

This section describes the details of building the robotic manipulation platform in simulation. We adopt Isaac Sim as the simulation environment and deploy both the Franka Emika Panda and Franka Research 3 robotic arms within it. For motion planning, we utilize Curobo as the solver, which computes feasible trajectories given the target end-effector poses. We use the default illumination settings in the simulation environment.

\begin{figure}[!htbp]
    \centering
    \includegraphics[width=0.92\linewidth]{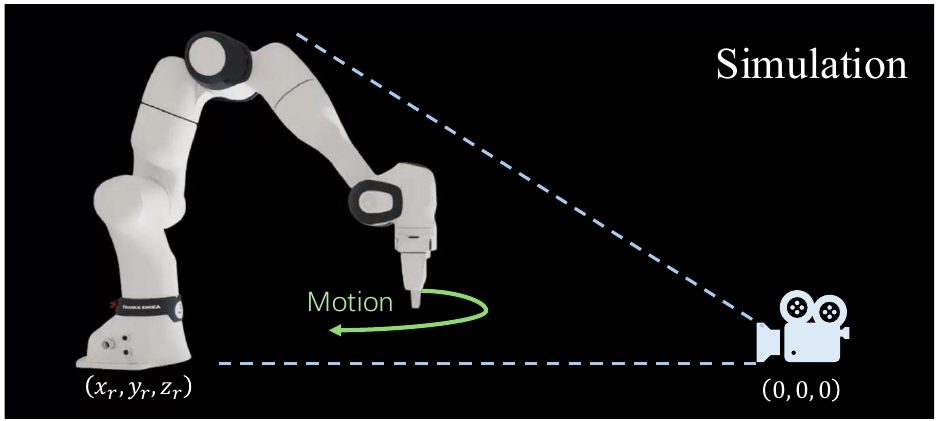}
    \caption{\textbf{Robot and Camera Placement in Simulation.} In simulation platforms such as IsaacSim, the virtual camera is placed at the position (0, 0, 0), while the robotic arm base is positioned at the corresponding point in the point cloud, denoted as $(x_r, y_r, z_r)$. RGB and depth data are collected during the robotic arm's motion.}
    \label{fig:fig_sup_2}
\end{figure}

As shown in Fig.~\ref{fig:fig_sup_2}, we place a virtual depth camera at the origin $\left[0, 0, 0\right]$ of the simulation scene, oriented relative to the robot’s base frame.
The camera’s focal length is set to match that used in the depth estimation module.
The robotic arm is positioned at a predefined spatial coordinate $\left[x_r, y_r, z_r\right]$ within the reconstructed point cloud space.
During robot motion, the camera operates at a sampling rate of 30~fps, capturing synchronized RGB and depth frames for subsequent reconstruction of the robot’s dynamic point cloud sequences.

\vspace{-0.3cm}

\section{Manipulation Synthesis Details}

We divide the point cloud sequence into three components: the background point cloud, the robot point cloud, and the object point cloud. Among them, the robot and object point clouds are dynamic, while the background point cloud remains static. We use GraspGen~\cite{murali2025graspgen} for grasp pose estimation. 

The grasp width is inferred from the inter-point distance along the principal axis of the reconstructed object point cloud. At the grasping moment $t_g$, the end-effector pose is denoted as $\mathbf{T}_{t_g}$ and the object pose as $\mathbf{T}_{\text{obj}, t_g}$. 
During the subsequent manipulation at time $t$, given the current end-effector pose $\mathbf{T}_t$, 
the object pose in the scene can be computed through rigid-body transformation as:
\begin{equation}
    \mathbf{T}_{\text{obj}, t} = \mathbf{T}_t \, \mathbf{T}_{t_g}^{-1} \, \mathbf{T}_{\text{obj}, t_g}.
    \label{eq:rigid_transform}
\end{equation}

\vspace{-0.3cm}
\begin{figure}[!htbp]
    \centering
    \includegraphics[width=0.85\linewidth]{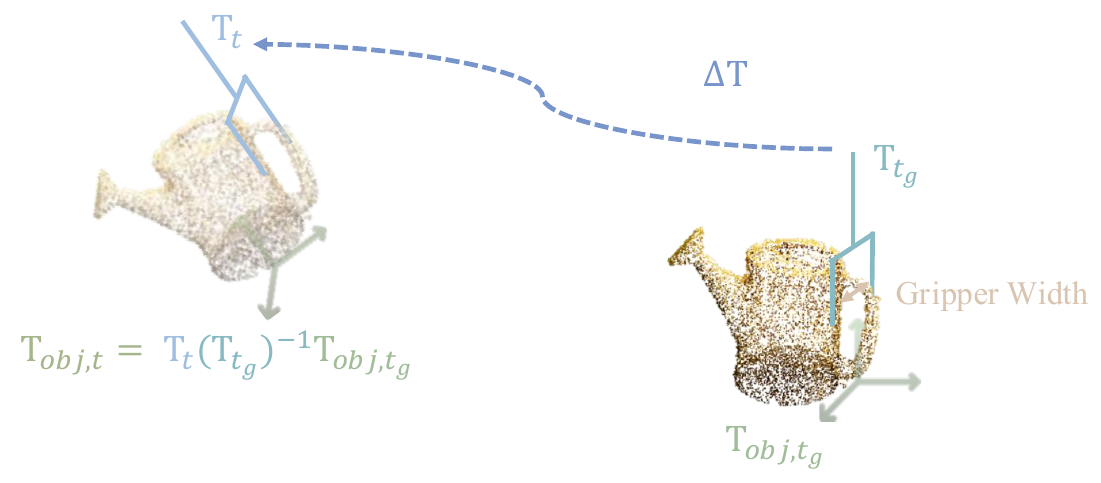}
    \caption{\textbf{Point Cloud Synthesis during Manipulation.} At time $t_g$, the object is grasped. The gripper width is calculated based on the point cloud, and the transformation of the object point cloud at time $t$ is computed according to the end-effector's pose.}
    \label{fig:fig_sup_3}
\end{figure}
\vspace{-0.3cm}

% FIG 5放这儿来了 还在调
\vspace{-0.5cm}
\begin{figure*}[!b]
    \centering
    \includegraphics[width=0.82\linewidth]{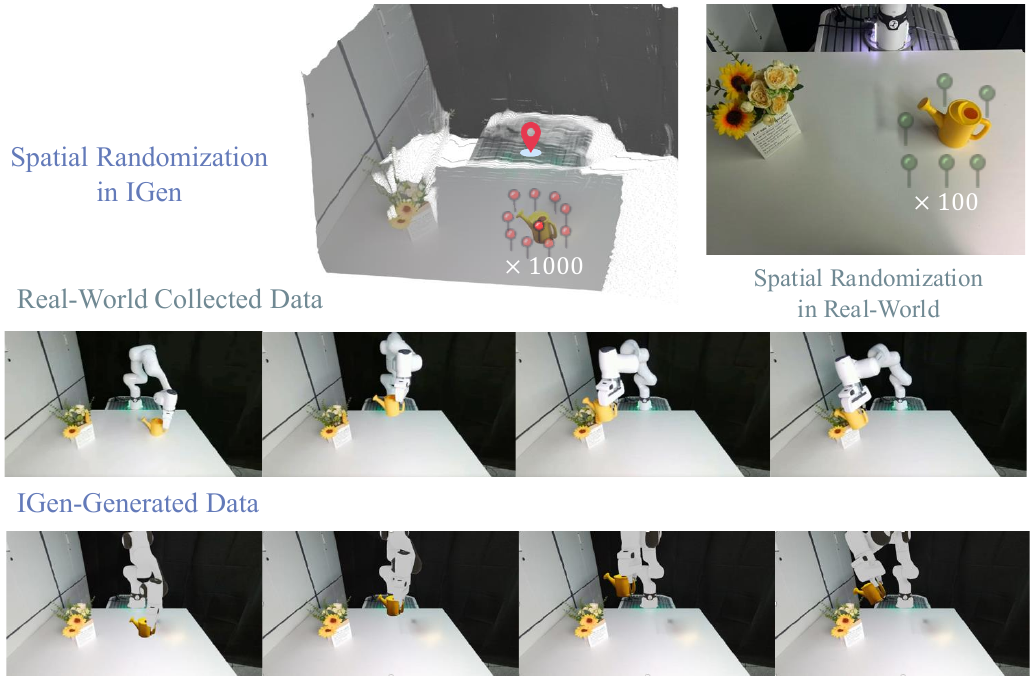}
    \caption{Spatial randomization of real-world data and IGen-generated data. The task is \textit{Grab the watering can and water the flowers.}
    }
    \label{fig:fig_sup_5_1}
\end{figure*}

\section{Real-World Experiment Details}

\begin{figure}[!htbp]
    \centering
    \includegraphics[width=0.92\linewidth]{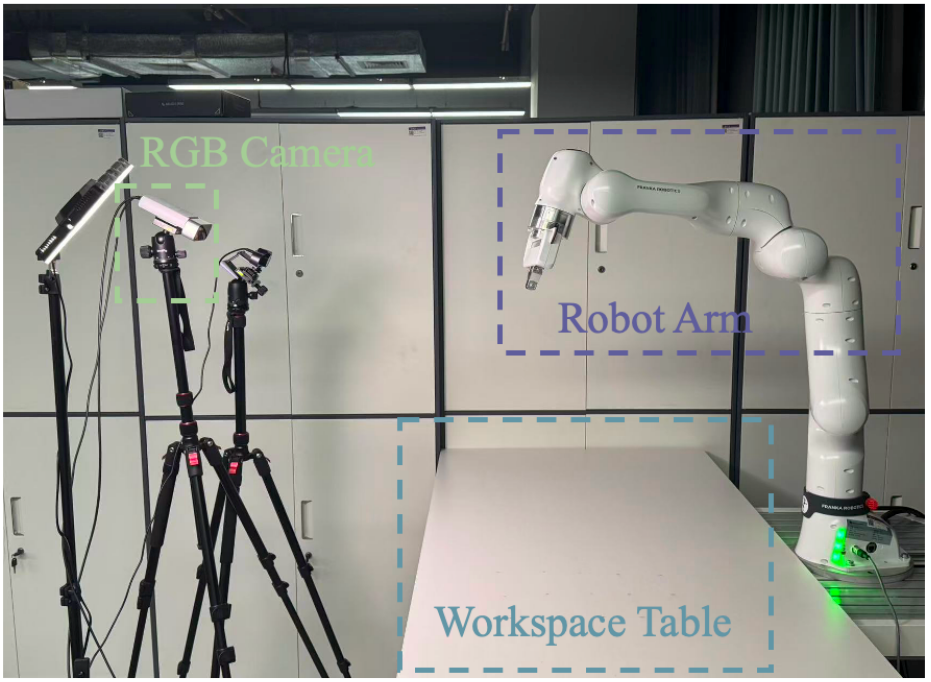}
    \caption{\textbf{Hardware Setup.} Our experimental setup consists of a Franka Research 3 robotic arm, a tabletop workspace, and a global RGB camera.}
    \label{fig:fig_sup_4}
\end{figure}

\vspace{1mm}\noindent\textbf{Hardware Setup.}~As shown in Fig.~\ref{fig:fig_sup_4}, we set up a real-world evaluation environment using the Franka Research 3 robotic arm. A Microsoft Kinect camera is placed in front of the robot to provide RGB visual input. The robotic arm performs task operations on the tabletop.

\begin{figure*} [!htbp]
    \centering
    \includegraphics[width=0.9\linewidth]{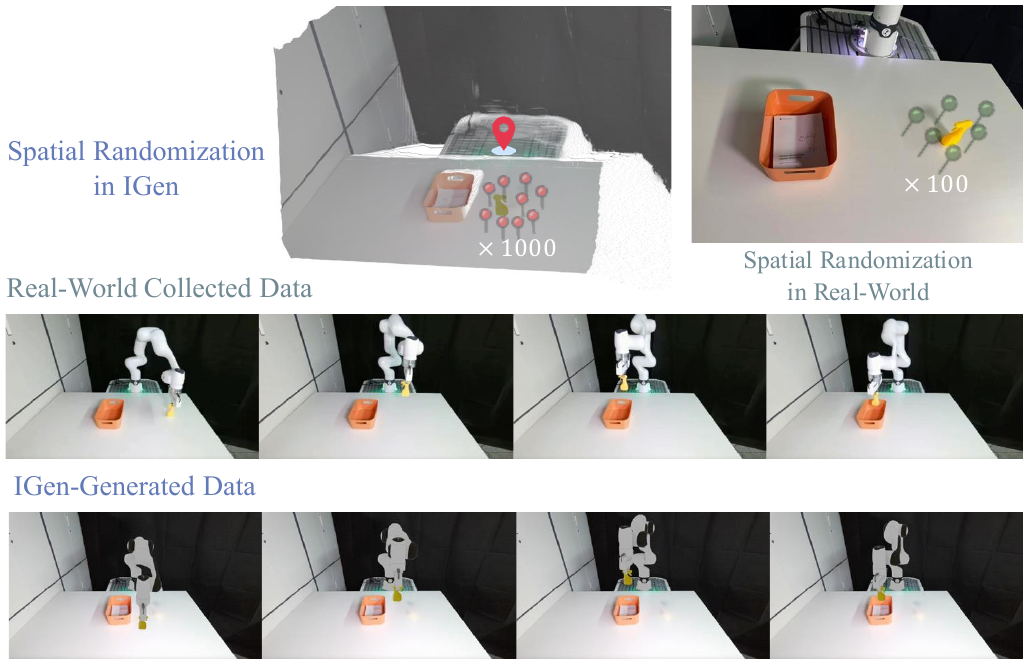}
    \caption{Spatial randomization of real-world data and IGen-generated data. The task is \textit{Pick up the bottle and place it into the basket.}
    }
    \label{fig:fig_sup_5_2}
\end{figure*}

\begin{figure*} [!htbp]
    \centering
    \includegraphics[width=0.9\linewidth]{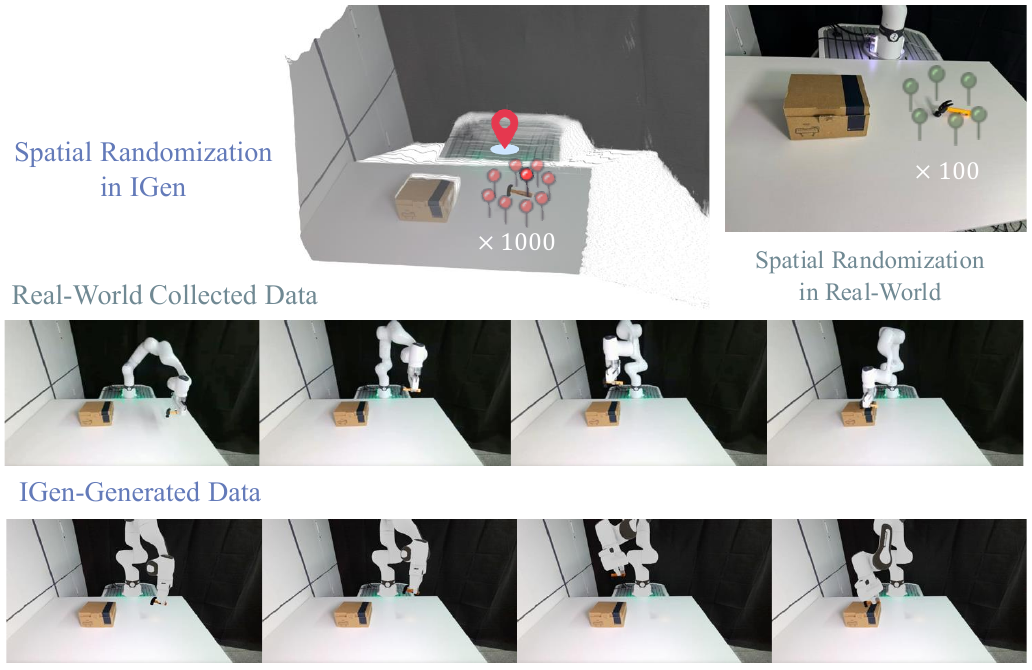}
    \caption{Spatial randomization of real-world data and IGen-generated data. The task is \textit{Use the hammer to hit the cardboard box.}
    }
    \label{fig:fig_sup_5_3}
\end{figure*}

\vspace{1mm}\noindent\textbf{Spatial Randomization.}~For real-world data collection, we sample random object positions within a $40~\text{cm} \times 30~\text{cm}$ tabletop grid. In IGen, spatial randomization is performed based on the point cloud of the placement area (e.g., the tabletop surface). We define a $200 \times 150$ pixel grid as the sampling region for randomization, ensuring that the spatial distribution closely matches that of the real-world setup. Regarding the spatial randomization of real-world data and IGen-generated data, see Fig.~\ref{fig:fig_sup_5_1}, \ref{fig:fig_sup_5_2}, and \ref{fig:fig_sup_5_3}.

\vspace{1mm}\noindent\textbf{Task Evaluation.}~We design diverse manipulation tasks involving complex interactions between objects and the surrounding scene. 
% The corresponding task instructions are as follows: \textit{(1) Pick up the yellow bottle and place it in the basket. (2) Lift the yellow watering can and water the flowers. (3) Lift the hammer and strike the cardboard box.}
For each task, we conduct 12 independent trials. Object initial positions are sampled on a $30~\text{cm} \times 25~\text{cm}$ tabletop grid, with a spacing of $7~\text{cm}$ between adjacent positions. All models are evaluated using the same set of initial object positions.

\begin{figure*} [!t]
    \centering
    \includegraphics[width=0.97\linewidth]{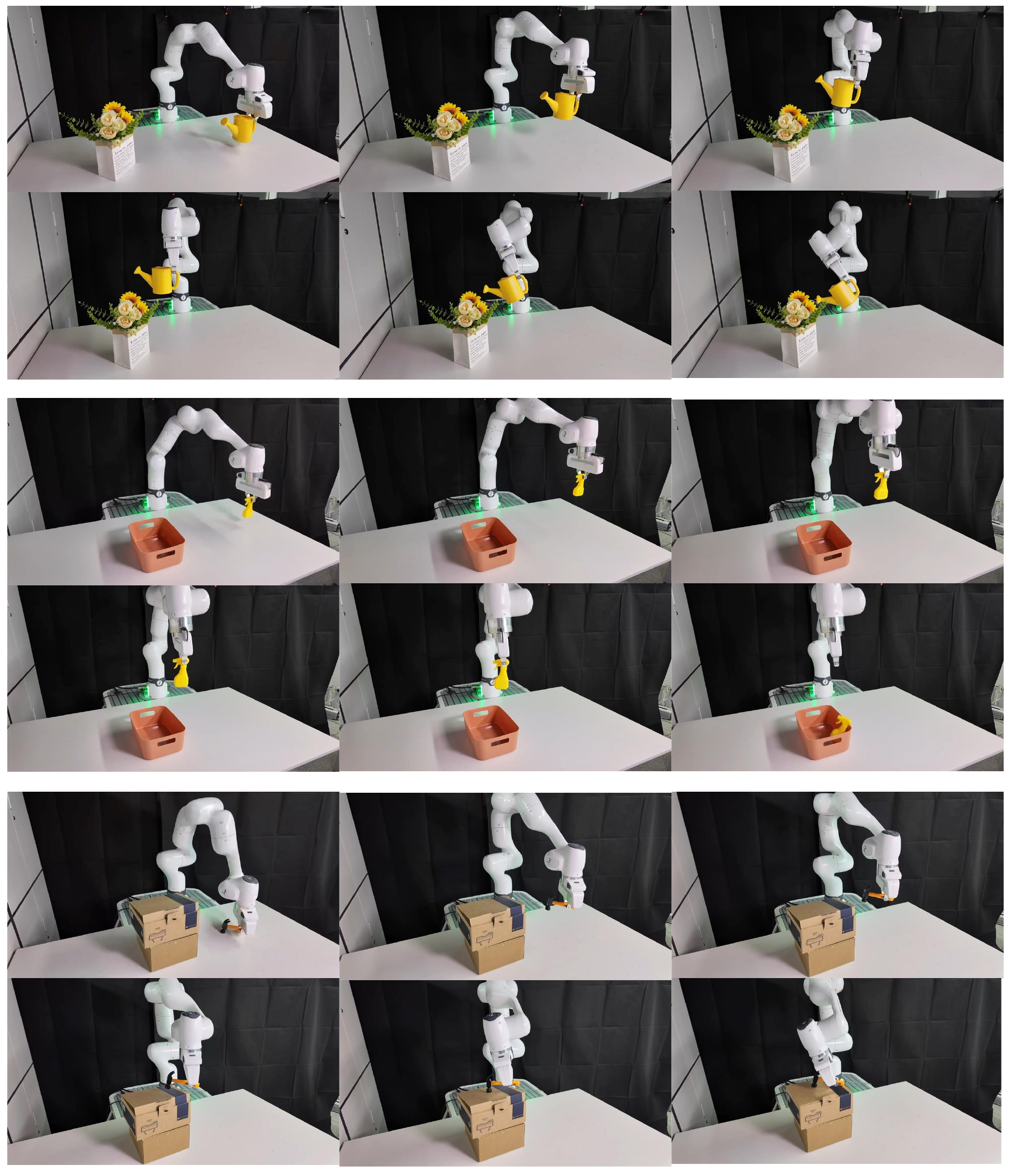}
    \caption{\textbf{Real-World Deployment of Policy trained with IGen-Generated Data.}~The task instructions are as follows:     \textit{Grab the watering can and water the flowers.} \textit{Pick up the bottle and place it into the basket.} \textit{Use the hammer to hit the cardboard box.}}
    \label{fig:fig_sup_6_1}
\end{figure*}

\begin{figure*} [!t]
    \centering
    \includegraphics[width=0.97\linewidth]{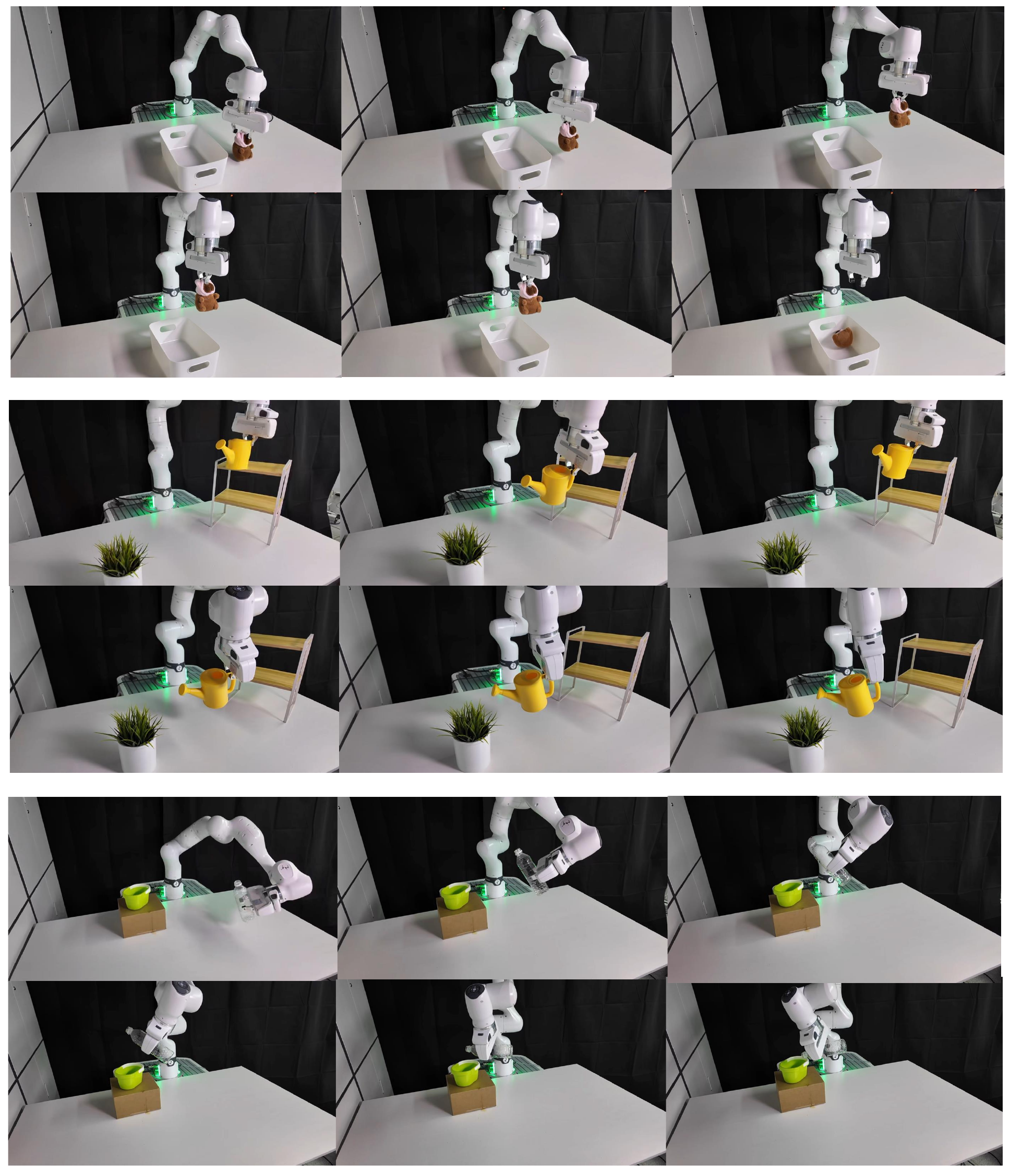}
    \caption{\textbf{Real-World Deployment of Policy trained with IGen-Generated Data.}~The task instructions are as follows:     \textit{Grasp the toy and put it into the bin.} \textit{Use the watering can on the cabinet to water the flowers.} \textit{Pour water from the plastic bottle into the container.}}
    \label{fig:fig_sup_6_2}
\end{figure*}

\vspace{1mm}\noindent\textbf{Policy Learning.}~This section describes the fine-tuning process of policy. We fine-tune ${\pi_{0}}$-base~\cite{black2024pi_0} for 30k training steps using LoRA~\cite{hu2022lora} with a batch size of 8. The model takes as input a single $224 \times 224$ RGB image and the absolute joint positions as the state, and predicts a 10-step relative joint angle action chunk. Training is conducted on a single NVIDIA A40 GPU, requiring approximately 10.8 hours per training. The performance of the model in real-world deployment is shown in Fig.~\ref{fig:fig_sup_6_1} and~\ref{fig:fig_sup_6_2}.

% \clearpage
\vspace{-0.5cm}
\vspace{-0.2cm}
\vspace{-0.2cm}

\vspace{-0.2cm}

\onecolumn
\section{Ablation Study}

\begin{figure*}  [!htbp]
    \vspace{-0.5cm}
    \centering
    \includegraphics[width=0.99\linewidth]{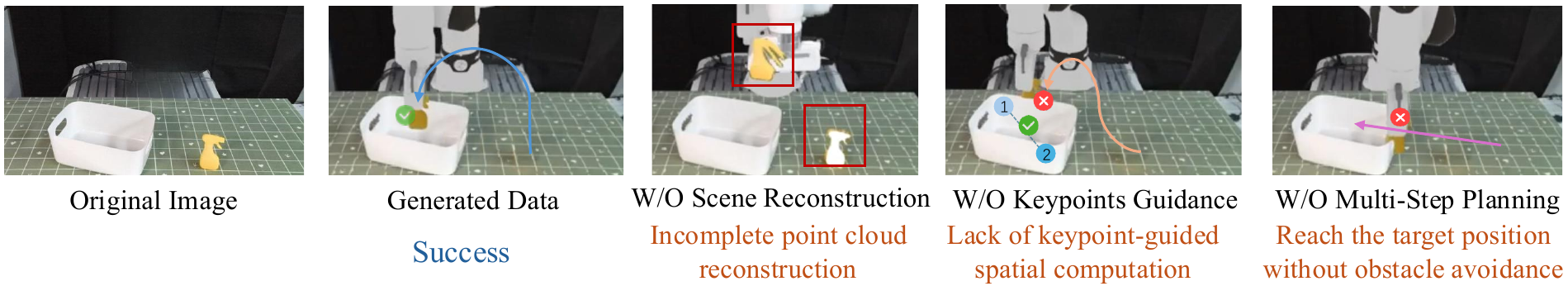}
    \vspace{-0.1cm}
    \caption{\textbf{Ablation study on different components of the pipeline.}~}
    \vspace{-0.5cm}
    \label{fig:fig_sup_ablation}
\end{figure*}

\section{Details for Keypoints Generation}

\begin{figure*}  [!htbp]
    \centering
    \vspace{-0.5cm}
    \includegraphics[width=0.99\linewidth]{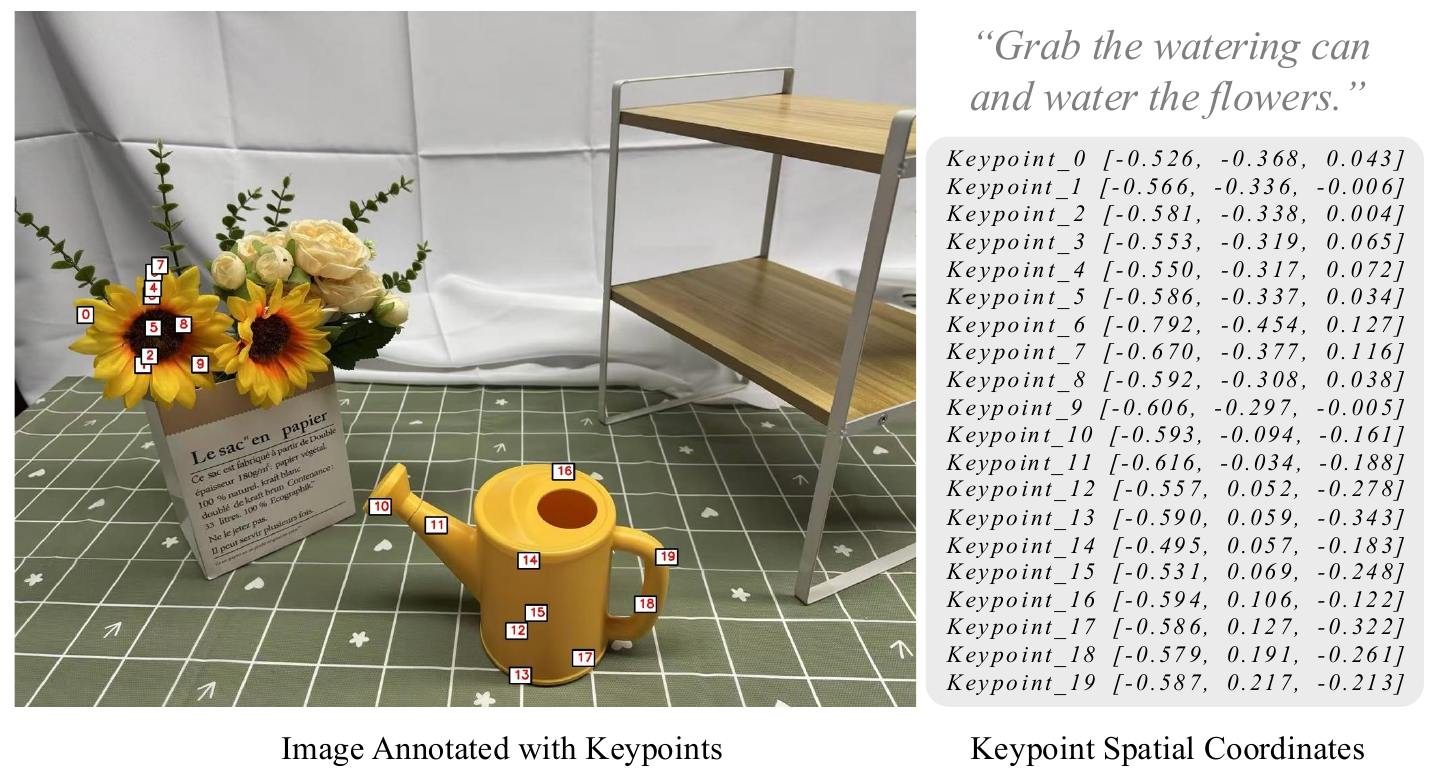}
    \vspace{-0.3cm}
    \caption{\textbf{Image with annotated keypoints, keypoint coordinates, and task description as input for VLM.}~}
    \vspace{-0.5cm}
    \label{fig:fig_sup_7}
\end{figure*}

% {
%     \small
%     \bibliographystyle{ieeenat_fullname}
%     \bibliography{main}
% }

\end{document}